\documentclass[11pt,a4paper]{article}
\usepackage{times,latexsym}
\usepackage{url}
\usepackage[T1]{fontenc}
\usepackage{enumitem}
\usepackage{graphicx}
\usepackage{todonotes}
\usepackage{booktabs}
\usepackage[noend]{algorithm2e}
\usepackage{amsmath,amssymb,mathtools,bm}
\usepackage{linguex}
\usepackage{comment}

\graphicspath{ {./plots/} }

\usepackage[acceptedWithA]{tacl2018v2}
\usepackage{xcolor,colortbl}

 \usepackage{array,multirow,graphicx}

\usepackage{xspace,mfirstuc,tabulary}

\newif\iftaclinstructions
\taclinstructionsfalse 
\iftaclinstructions
\renewcommand{\confidential}{}
\renewcommand{\anonsubtext}{(No author info supplied here, for consistency with
TACL-submission anonymization requirements)}
\newcommand{\instr}
\fi

\iftaclpubformat 

\else

\fi

\usepackage{cleveref}

\renewcommand{\autoref}[1]{\Cref{#1}}

\title{Decomposing and Recomposing Event Structure}

\author{
 William Gantt \\
 University of Rochester \And
 Lelia Glass \\
 Georgia Institute of Technology \And
 Aaron Steven White \\
 University of Rochester
}

\date{}

\begin{document}
\maketitle
\begin{abstract}
  We present an event structure classification empirically derived from inferential properties annotated on sentence- and document-level Universal Decompositional Semantics (UDS) graphs. We induce this classification jointly with semantic role, entity, and event-event relation classifications using a document-level generative model structured by these graphs. To support this induction, we augment existing annotations found in the UDS1.0 dataset, which covers the entirety of the English Web Treebank, with an array of inferential properties capturing fine-grained aspects of the temporal and aspectual structure of events. The resulting dataset (available at \href{http://decomp.io}{decomp.io}) is the largest annotation of event structure and (partial) event coreference to date. 
\end{abstract}

\setlength{\Exlabelsep}{.3em}
\setlength{\Extopsep}{.05\baselineskip}

\section{Introduction}
\label{sec:introduction}

Natural language provides myriad ways of communicating about complex events. For instance,
one and the same event can be described at a
coarse grain, using a single clause \ref{ex:build-coarse}, or at a finer grain, using an entire document \ref{ex:build-fine}.

\ex. The contractors built the house. \label{ex:build-coarse}

\ex. They started by laying the house's foundation. They then framed the house before installing the plumbing. After that[...]\label{ex:build-fine}

Further, descriptions of the same event at different granularities can be interleaved within the same document---e.g. \ref{ex:build-fine} might well directly follow \ref{ex:build-coarse} as an elaboration on the house-building process. 

Consequently, extracting knowledge about complex events from text involves determining the structure of the events being referred to: what their parts are, how those parts are laid out in time, who participates in them and how, etc. Determining this structure requires an event classification whose elements are associated with event structure representations. A number of such classifications and annotated corpora exist: FrameNet \citep{baker_berkeley_1998}, VerbNet \citep{KipperSchuler:2005}, PropBank \citep{palmer_proposition_2005}, Abstract Meaning Representation \citep[][]{banarescu-etal-2013-abstract}, and Universal Conceptual Cognitive Annotation \citep[][]{abend-rappoport-2013-universal} among others. 

Similar in spirit to this prior work, but different in method, our work aims to develop an \textit{empirically derived} event structure classification. Where prior work takes a top-down approach---hand-engineering an event classification before deploying it for annotation---we take a bottom-up approach---\textit{decomposing} event structure into a wide variety of theoretically informed, cross-cutting semantic properties, annotating for those properties, then \textit{recomposing} an event classification from them by induction. The properties on which our categories rest target (i) the substructure of an event---e.g. that the building described in \ref{ex:build-coarse} consists of a sequence of subevents resulting in the creation of some artifact; (ii) the superstructure in which an event takes part---e.g. that laying a house's foundation is part of building a house, alongside framing the house, installing the plumbing, etc.; (iii) the relationship between an event and its participants---e.g. that the contractors in \ref{ex:build-coarse} 
build the house collectively through their joint efforts; and
(iv) properties of the event's participants---e.g. that the contractors in \ref{ex:build-coarse} are animate while the house is not. 

To derive our event structure classification, we extend the Universal Decompositional Semantics dataset \citep[UDS;][]{white-etal-2016-universal,white-etal-2020-universal}. UDS annotates for a subset of key event structure properties, but a range of key properties remain to be captured. After motivating the need for these additional properties (\S \ref{sec:background}), we develop annotation protocols for them (\S \ref{sec:annotation-protocol}). We validate our protocols (\S\ref{sec:validation-experiments}) and use them to collect annotations for the entire Universal Dependencies \citep{nivre-etal-2016-universal} English Web Treebank (\S\ref{sec:corpus-annotation}; \citealt{Bies-etal:2012}), resulting in the UDS-EventStructure dataset (UDS-E). To derive an event structure classification from UDS-E and existing UDS annotations, we develop a document-level generative model that jointly induces event, entity, semantic role, and event-event relation types (\S\ref{sec:event-structure-induction}). Finally, we compare these types to those found in existing event structure classifications (\S\ref{sec:comparison-to-existing-ontologies}). We make UDS-E and our code available at \href{http://decomp.io}{decomp.io}.

\section{Background}
\label{sec:background}

Contemporary theoretical treatments of event structure tend to take as their starting point \citeauthor{vendler_verbs_1957}'s (\citeyear{vendler_verbs_1957}) seminal four-way classification. We briefly discuss this classification and elaborations thereon before turning to other event structure classifications developed for annotating corpora.\footnote{The theoretical literature on event structure is truly vast. See \citealt{truswell_oxford_2019} for a collection of overview articles.} We then contrast these with the fully decompositional approach we take in this paper.


\vspace{-2mm}

\paragraph{Theoretical Approaches}

\citeauthor{vendler_verbs_1957} categorizes event descriptions into four classes: \textit{statives} \ref{ex:state},  \textit{activities} \ref{ex:activity}, \textit{achievements} \ref{ex:achievement}, and \textit{accomplishments} \ref{ex:accomplishment}. As theoretical constructs, these classes are used to explain both the distributional characteristics of event descriptions as well as inferences about how an event progresses over time.

\ex. \label{ex:state}Jo was in the park.\\\phantom{---------------}\textit{stative} = [$+$\textsc{dur}, $-$\textsc{dyn}, $-$\textsc{tel}]

\ex. \label{ex:activity}Jo ran around in the park.\\\phantom{--------------}\textit{activity} = [$+$\textsc{dur}, $+$\textsc{dyn}, $-$\textsc{tel}]

\ex. \label{ex:achievement}Jo arrived at the park.\\\phantom{--------}\textit{achievement} = [$-$\textsc{dur}, $+$\textsc{dyn}, $+$\textsc{tel}]

\ex. \label{ex:accomplishment}Jo ran to the park.\\\phantom{---}\textit{accomplishment} = [$+$\textsc{dur}, $+$\textsc{dyn}, $+$\textsc{tel}]

Work building on \citeauthor{vendler_verbs_1957}'s discovered that these classes can be decomposed into the now well-accepted component properties in \ref{ex:durativity}--\ref{ex:telicity} \citep[][]{kenny_action_1963,lakoff_nature_1965,verkuyl_compositional_1972,bennett_towards_1978,mourelatos_events_1978,dowty_word_1979}. 

\ex. \textsc{dur(ativity)}: whether the event happens at an instant or extends over time \label{ex:durativity}

\ex. \textsc{dyn(amicity)}: whether the event involves change, broadly construed \label{ex:dynamicity}

\ex. \textsc{tel(icity)}: whether the event culminates in a participant changing state or location, being created or destroyed, etc. \label{ex:telicity}

Later work further expanded these properties and, therefore, the possible classes. Expanding on \textsc{dyn}, \citet{taylor_tense_1977} suggests a distinction between dynamic predicates that refer to events with dynamic subevents---e.g. the individual strides in a running---and ones that do not---e.g. the gliding in \ref{ex:no-natural-parts} \citep[see also][]{bach_algebra_1986,smith_modes_2003}.

\ex. The pelican glided through the air. \label{ex:no-natural-parts}

Dynamic events with dynamic subevents can be further distinguished based on whether the subevents are similar---e.g. the strides in a running---or dissimilar---e.g. the subevents in a house-building \citep{pinon_ontology_1995}. In the case where the subevents are similar and a participant itself has subparts---e.g. when the participant is a group---there may be a bijection from participant subparts to subevents. In \ref{ex:distributive}, there is a smiling for each child that makes up the composite smiling---\textit{smile} is \textit{distributive}. In \ref{ex:collective}, the meeting presumably has some structure, but there is no bijection from members to subevents---\textit{meet} is \textit{collective} \citep[see][for a review]{Champollion:2010}.  

\ex. \{The children, Jo and Bo\} smiled.\label{ex:distributive}

\ex. \{The committee, Jo and Bo\} met.\label{ex:collective}

Expanding on \textsc{tel}, \citet{dowty_thematic_1991} argues for a distinction among telics in which the culmination comes about incrementally \ref{ex:incremental} or abruptly \ref{ex:nonincremental} \citep[see also][]{tenny_grammaticalizing_1987,krifka_nominal_1989,krifka_thematic_1992,krifka_origins_1998,levin1991wiping,rappaport1998building,hovav2001,croft2012verbs}. 

\ex. The gardener mowed the lawn. \label{ex:incremental}

\ex. The climber summitted at 5pm. \label{ex:nonincremental}

This notion of incrementality is intimately tied up with the notion of \textsc{dur(ativity)}. For instance, \citet{moens_temporal_1988} point out that certain event structures can be systematically transformed into others---e.g. whereas \ref{ex:nonincremental} describes the summitting as something that happens at an instant (and is thus abrupt), \ref{ex:coerced} describes it as a process that culminates in having reached the top of the mountain \citep[see also][]{pustejovsky_generative_1995}.

\ex. The climber was summitting. \label{ex:coerced}

Such cases of \textit{aspectual coercion} highlight the importance of grammatical factors in determining the structure of an event. More general contextual factors are also at play when determining event structure: \textit{I ran} can describe a telic event---e.g. when it is known that I run the same distance or to the same place every day---or an atelic event---e.g. when the destination and/or distance is irrelevant in context \citep{dowty_word_1979,Olsen:1997}. This context-sensitivity strongly suggests that annotating event structure is not simply a matter of building a type-level lexical resource and projecting its labels onto text: actual text must be annotated.

\vspace{-2mm}

\paragraph{Resources}

Early, broad-coverage lexical resources, such as the Lexical Conceptual Structure lexicon \citep[LCS;][]{dorr_machine_1993}, attempt to directly encode an elaboration of the core Vendler classes in terms of a hand-engineered graph representation proposed by \citet{jackendoff_semantic_1990}. VerbNet \citep{KipperSchuler:2005} further elaborates on LCS by building on the fine-grained syntax-based classification of \citet{levin_english_1993} and links her classes to LCS-like representations. More recent versions of VerbNet \citep[v3.3+;][]{brown-etal-2018-integrating} update these representations to ones based on the Dynamic Event Model \citep[][]{pustejovsky-2013-dynamic}. 

COLLIE-V, which expands the TRIPS lexicon and ontology (\citealt{ferguson1998trips} \textit{et seq}), takes a similar tack of producing hand-engineered event structures, combining this hand-engineering with a procedure for bootstrapping event structures \citep{allen-etal-2020-broad}. FrameNet also contains hand-engineered event structures, though they are significantly more fine-grained than those found in LCS or VerbNet \citep{baker_berkeley_1998}.

VerbNet, COLLIE-V, and FrameNet are not directly annotated on text, though annotations for at least VerbNet and FrameNet can be obtained by using SemLink to project FrameNet and VerbNet annotations onto PropBank annotations \citep{palmer_proposition_2005}. PropBank frames have been enriched in a variety of other ways. One such enrichment can be found in Abstract Meaning Representation \citep[AMR;][]{banarescu-etal-2013-abstract,donatelli2018annotation}. Another can be found in Richer Event Descriptions \citep[RED;][]{ogorman-etal-2016-richer}, which annotates events and entities for factuality (whether an event actually happened) and genericity (whether an event/entity is a particular or generic) as well as annotating for causal, temporal, sub-event, and co-reference relations between events \citep[see also][]{chklovski2004verbocean,hovy2013events,cybulska-vossen-2014-using}.

\vspace{-1mm}

Additional less fine-grained event classifications exist in TimeBank \citep{pustejovsky_timebank_2006}, Universal Conceptual Cognitive Annotation \citep[UCCA;][]{abend-rappoport-2013-universal}, and the Situation Entities dataset \citep[SitEnt;][]{friedrich-palmer-2014-situation,friedrich-etal-2016-situation}. Of these, the closest to capturing the standard Vendler classification and decompositions thereof is SitEnt. The original version of SitEnt annotates only for a state-event distinction (alongside related, non-event structural distinctions), but later elaborations further annotate for telicity \citep{friedrich-gateva-2017-classification}. Because of this close alignment to the standard Vendler classes, we use SitEnt annotations as part of validating our own annotation protocol in \S\ref{sec:annotation-protocol}.

\vspace{-2mm}

\paragraph{Universal Decompositional Semantics}

In contrast to the hand-engineered event structure classifications discussed above, our aim is to derive event structure representations directly from semantic annotations. To do this,  we extend the existing annotations in the Universal Decompositional Semantics dataset (UDS; \citealt{white-etal-2016-universal, white-etal-2020-universal}) with key annotations for the event structural distinctions discussed above. Our aim is not necessarily to reconstruct any previous classification, though we do find in \S\ref{sec:event-structure-induction} that our event type classification approximates Vendler's to some extent.

UDS is a semantic annotation framework and dataset based on the principles that (i) the semantics of words or phrases can be \textit{decomposed} into sets of simpler semantic properties and (ii) these properties can be annotated by asking straightforward questions intelligible to non-experts. UDS comprises two layers of annotations on top of the Universal Dependencies (UD) syntactic graphs in the English Web Treebank (EWT): (i) predicate-argument graphs with mappings into the syntactic graphs, derived using the PredPatt tool \citep{white-etal-2016-universal, zhang_evaluation_2017}; and (ii) crowd-sourced annotations for properties of events (on the \textit{predicate nodes} of the predicate-argument graph), entities (on the \textit{argument nodes}), and their relationship (on the \textit{predicate-argument edges}). 

The UDS properties are organized into three \textit{predicate subspaces} with five properties in total:

\begin{figure*}
    \centering
    \includegraphics[width=1.7\columnwidth]{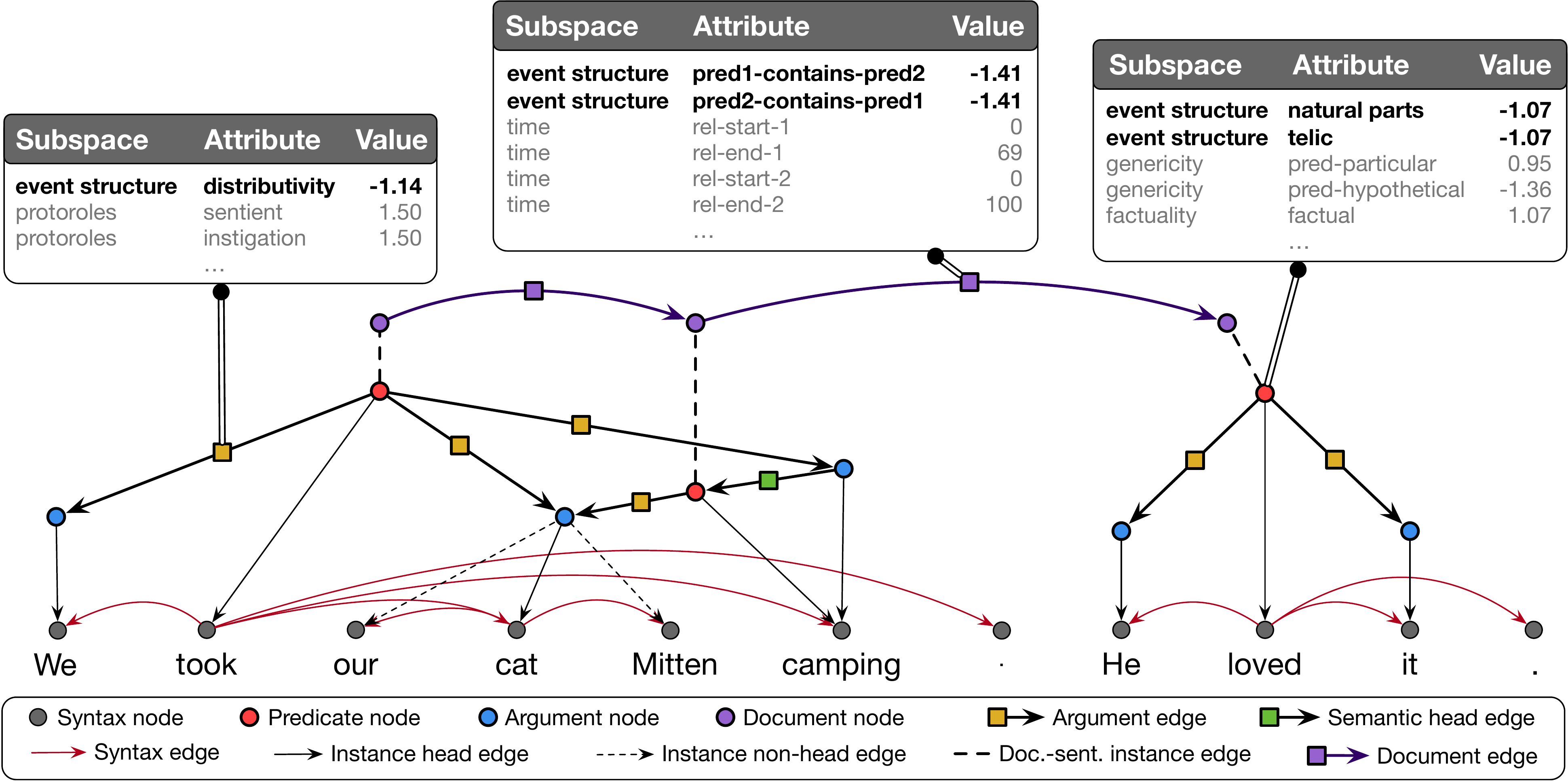}
    \vspace{-3mm}
    \caption{Example UDS semantics and syntax graphs with select properties (see \autoref{fn:property-mem} on the property values). Bolded properties are ones we collect in this paper, and our new \textit{document-level} graph is also shown in purple.}
    \vspace{-5mm}
    \label{fig:uds_semantics_graph}
\end{figure*}

\vspace{-3mm}

\begin{itemize}[itemsep=-2pt]
    \item \textsc{factuality} \citep{rudinger-etal-2018-neural}\\\textit{factual}: did the event happen?
    \item \textsc{genericity} \citep{govindarajan-etal-2019-decomposing}\\
          \textit{kind}: is the event generic?\\
          \textit{hypothetical}: is the event hypothetical?\\
          \textit{dynamic}: is the event dynamic or stative?
    \item \textsc{time} \citep{vashishtha-etal-2019-fine}\\
          \textit{duration}: how long did/will the event last?
\end{itemize}

\vspace{-3mm}

\noindent Two \textit{argument subspaces} with four properties:

\vspace{-3mm}

\begin{itemize}[itemsep=-2pt]
    \item \textsc{genericity} \citep{govindarajan-etal-2019-decomposing}\\
          \textit{particular}: is the entity a particular?\\
          \textit{kind}: is the entity a kind?\\
          \textit{abstract}: is the entity abstract or concrete?
    \item \textsc{wordsense} \citep{white-etal-2016-universal}\\
          Which coarse entity types (WordNet supersense) does the entity have?
\end{itemize}

\vspace{-3mm}

\noindent And one \textit{predicate-argument subspace} with 16 properties (see \citealt{white-etal-2016-universal} for full list):

\vspace{-3mm}

\begin{itemize}[itemsep=-2pt]
    \item \textsc{protoroles} \citep{reisinger-etal-2015-semantic}\\
          \textit{instigation}: did participant cause event?\\
          \textit{change of state}: did participant change state during or as a consequence of event?\\
          \textit{change of location}: did participant change location during event?\\
          \textit{existed \{before, during, after\}} did participant exist \{before, during, after\} the event?
\end{itemize}

\vspace{-3mm}

\noindent \autoref{fig:uds_semantics_graph} shows an example UDS1.0 graph \citep{white-etal-2020-universal} augmented with (i) a subset of the properties we add in bold (see \S\ref{sec:annotation-protocol}); and (ii) \textit{document-level} edges in purple (see \S\ref{sec:event-structure-induction}).\footnote{Following \citet{white-etal-2020-universal}, the property values in \autoref{fig:uds_semantics_graph} are derived from raw annotations using mixed effects models \citep[MEMs;][]{gelman_data_2014}, which enable one to adjust for differences in how annotators approach a particular annotation task \citep[see also][]{gantt-etal-2020-natural}. In \S\ref{sec:event-structure-induction}, we similarly use MEMs in our event structure induction model, allowing us to work directly with the raw annotations. \label{fn:property-mem}}

The UDS annotations and associated toolkit have supported research in a variety of areas, including syntactic and semantic parsing \citep{stengel-eskin-etal-2020-universal,stengel2021joint}, semantic role labeling \citep{teichert2017semantic} and induction \citep{white-etal-2017-semantic}, event factuality prediction \citep{rudinger-etal-2018-neural}, temporal relation extraction \citep{vashishtha-etal-2019-fine}, among others. For our purposes, the annotations do cover some event structural distinctions---e.g.\ \textit{dynamicity}, specific cases of \textit{telicity} (in the form of \textit{change of state}, \textit{change of location}, and \textit{existed \{before, during, after\}}), and \textit{durativity}. In this sense, UDS provides an alternative, decompositional event representation that distinguishes it from more traditional categorical ones like SitEnt. However, the existing annotations fail to capture a number of the core distinctions above---a lacuna this work aims to fill.

\section{Annotation Protocol}
\label{sec:annotation-protocol}

We annotate for the core event structural distinction not currently covered by UDS, breaking our annotation into three subprotocols.  For all questions, annotators report confidence in their response to each question on a scale from 1 (\textit{not at all confident}) to 5 (\textit{totally confident}).\footnote{The annotation interfaces for all three subprotocols, including instructions, are available at \href{http://decomp.io}{decomp.io}.}

\vspace{-2mm}

\paragraph{Event-subevent}

Annotators are presented with a sentence containing a single highlighted predicate followed by four questions about the internal structure of the event it describes. Q1 asks whether the event described by the highlighted predicate has natural subparts. Q2 asks whether the event has a natural endpoint. 

The final questions depend on the response to Q1. If an annotator responds that the highlighted predicate refers to an event that \textit{has} natural parts, they are asked (i) whether the parts are similar to one another and (ii) how long each part lasts on average. If an annotator instead responds that the event referred to does \textit{not} have natural parts, they are asked (i) whether the event is dynamic, and (ii) how long the event lasts.

All questions are binary except those concerning duration, for which answers are supplied as one of twelve ordinal values \citep[see][]{vashishtha-etal-2019-fine}: \textit{effectively no time at all, fractions of a second, seconds, minutes, hours, days, weeks, months, years, decades, centuries} or \textit{effectively forever}. Together, these questions target the three Vendler-inspired features (\textsc{dyn}, \textsc{dur}, \textsc{tel}), plus a fourth dimension for subtypes of dynamic predicates. In the context of UDS, these properties form a predicate node subspace, alongside \textsc{factuality}, \textsc{genericity}, and \textsc{time}.


\vspace{-2mm}

\paragraph{Event-event}

Annotators are presented with either a single sentence or a pair of adjacent sentences, with the two predicates of interest highlighted in distinct colors. For a predicate pair $(p_1, p_2)$ describing an event pair $(e_1, e_2)$, annotators are asked whether $e_1$ is a mereological part of $e_2$, and vice versa. Both questions are binary: a positive response to both indicates that $e_1$ and $e_2$ are the same event; and a positive response to exactly one of the questions indicates proper parthood. 
Prior versions of UDS do not contain any predicate-predicate edge subspaces, so we add document-level graphs to UDS (\S\ref{sec:event-structure-induction}) to capture the relation between adjacently described events.


This subprotocol targets generalized event coreference, identifying \textit{constituency} in addition to strict identity. It also augments the information collected in the event-subevent protocol: insofar as a proper subevent relation holds between $e_1$ and $e_2$, we obtain additional fine-grained information about the subevents of the containing event---e.g. an explicit description of at least one subevent.

\vspace{-2mm}

\paragraph{Event-entity}

The final subprotocol focuses on the relation between the event described by a predicate and its plural or conjoined arguments, asking whether the predicate is distributive or collective with respect to that argument. This property accordingly forms a predicate-argument subspace in UDS, similar to \textsc{protoroles}.



\section{Validation Experiments}
\label{sec:validation-experiments}

We validate our annotation protocol (i) by assessing interannotator agreement (IAA) among both experts and crowd-sourced annotators for each subprotocol on a small sample of items drawn from existing annotated corpora (\S\ref{ssec:item-selection}-\ref{ssec:interannotator-agreement}); and (ii) by comparing annotations generated using our protocol against existing annotations that cover (a subset of) the phenomena that ours does and are generated by highly trained annotators (\S\ref{ssec:protocol-comparison}).

\subsection{Item Selection}
\label{ssec:item-selection}

For each of the three subprotocols, one of the authors selected 100 sentences for inclusion in the pilot for that subprotocol. This author did not consult with the other authors on their selection, so that annotation could be blind. 

For the event-subevent subprotocol, the 100 sentences come from the portion of the MASC corpus \citep{ide_masc_2008} that \citet{friedrich-etal-2016-situation} annotate for eventivity (\textsc{event} v. \textsc{state}) and that \citet{friedrich-gateva-2017-classification} annotate for telicity (\textsc{telic} v. \textsc{atelic}). For the event-event subprotocol, the 100 sentences come from the portions of the Richer Event Descriptions corpus \citep[RED;][]{ogorman-etal-2016-richer} that are annotated for event subpart relations. To our knowledge, no existing annotations cover distributivity, and so for our event-entity protocol, we select 100 sentences (distinct from those used for the event-subevent subprotocol) and compute IAA, but do not compare against existing annotations.

\begin{table*}[t]
\footnotesize
    \centering
\begin{tabular}{p{0.2cm} | p{3cm} | p{2.5cm} | p{8.5cm} }
    
		&   \textbf{Annotation}   & \textbf{Count (\%)} & \textbf{Example} \\ \hline
 \parbox[t]{2mm}{\multirow{11}{*}{\rotatebox[origin=c]{90}{\textit{Event-subevent}}}} 
		&  Has natural parts 			& 6,903 (23\%) 			& The eighteen steps of the dance are \underline{done} rhythmically  \\ 
		& \cellcolor{lightgray} \hspace{0.3cm} Parts  similar		 & \cellcolor{lightgray} \hspace{0.3cm} 4,498 (15\%) 			&  \cellcolor{lightgray} \hspace{0.3cm} Israel resumed its policy of \underline{targeting} militant leaders  \\ 
		&  \cellcolor{lightgray}   \hspace{0.3cm} Parts dissimilar &  \cellcolor{lightgray} \hspace{0.3cm} 2,158  (7\%) 			& \cellcolor{lightgray}  \hspace{0.3cm} Fish are probably the easiest to \underline{take} care of  \\ 
		& \hspace{0.3cm} \cellcolor{lightgray}  (Part duration) 		& \cellcolor{lightgray}  \hspace{0.3cm}  (--)  			& \cellcolor{lightgray}  \hspace{0.3cm} (ordinal; not shown)  \\
		 &  No natural parts			 & 23,069 (77\%) 				& It \underline{had} better nutritional value \\ 
		& \hspace{0.3cm} \cellcolor{lightgray} Dynamic & \cellcolor{lightgray}  \hspace{0.1cm} 13,903 (48\%) & \cellcolor{lightgray} \hspace{0.3cm} I would like to informally \underline{get} together with you   \\
		&\hspace{0.3cm} \cellcolor{lightgray} Not dynamic &  \cellcolor{lightgray}  \hspace{0.3cm} 8,839 (29\%) & \cellcolor{lightgray} \hspace{0.3cm} I assume this \underline{is 12:30} Central Time? \\ 
		& \hspace{0.3cm} \cellcolor{lightgray} (Full duration) & \cellcolor{lightgray}  \hspace{0.3cm}  (--)  & \cellcolor{lightgray}  \hspace{0.3cm}  (ordinal; not shown)  \\
		&   Natural endpoint &   6,031 (20\%)  & I will \underline{deliver} it to you  \\ 
		&   No natural endpoint &  23,941 (80\%)  & If you \underline{know} or work there could you enlighten me? \\
		&  \cellcolor{lightgray} total &  \cellcolor{lightgray}  29,984 & \cellcolor{lightgray} (all event descriptions) \\ \hline
 \parbox[t]{2mm}{\multirow{5}{*}{\rotatebox[origin=c]{90}{\textit{Event-event}}}}
 		 &   P1, P2 identical & 2,435 (6\%) 						& All horses [\ldots]  are \underline{happy}$_{1}$ \& \underline{healthy}$_{2}$ when they arrive  \\
		& P1, P2 disjoint  & 30,247  (80\%) 						&   I am often \underline{stopped}$_{1}$ on the street and asked, `Who does your hair \ldots I \underline{LOVE}$_{2}$ it'  \\
		&   P1 $\subset$ P2 & 1,832  (5\%)						& The office is shared with a foot doctor and it's \underline{very sterile}$_{1}$ and medical \underline{feeling}$_{2}$, which I liked  \\
		& P2 $\subset$ P1  &  3,029 (8\%) 						&  It \underline{is a very cruel death}$_{1}$ with bodies \underline{dismembered}$_{2}$ \\ 
		& \cellcolor{lightgray}  total & \cellcolor{lightgray}  37,719	&\cellcolor{lightgray}  (pairs of event descriptions w/ temporal overlap) \\ \hline
\parbox[t]{2mm}{\multirow{3}{*}{\rotatebox[origin=c]{90}{\textit{Event-entity}}}}  
		&  Distributive & 4,812 (50\%) 						& the \textbf{pics} turned out \underline{ok}  \\
		& Collective  & 4,876 (50\%)  						& \textbf{we} \underline{draw} on our many faith traditions \ to arrive at a common  \newline conviction \\ 
&\cellcolor{lightgray}  total & \cellcolor{lightgray}   9,710 & \cellcolor{lightgray}  (event descriptions  with plural arguments) \\ \hline

\end{tabular}
\vspace{-4mm}
    \caption{Descriptive statistics and examples from Train and Dev data.  Each item was annotated by a single annotator in Train; and by three annotators in Dev, of which this table reports the majority opinion.}
    \label{tab:corpus-examples}
    \vspace{-5mm}
\end{table*}

\subsection{Interannotator Agreement}
\label{ssec:interannotator-agreement}

We compute two forms of IAA: (i) IAA among expert annotators (the three authors); and (ii) IAA between experts and crowd-sourced annotators. In both cases, we use Krippendorff's $\alpha$ as our measure of (dis)agreement \citep{krippendorff_content_2004}. For the binary responses, we use the nominal form of $\alpha$; for the ordinal responses, we use the ordinal.

\vspace{-2mm}

\paragraph{Expert Annotators}

For each subprotocol, the three authors independently annotated the 100 sentences selected for that subprotocol.

Prior to analysis, we ridit score the confidence ratings by annotator to normalize them for differences in annotator scale use (see \citealt{govindarajan-etal-2019-decomposing} for discussion of ridit scoring confidence ratings in a similar annotation protocol). This method maps ordinal labels to (0, 1) on the basis of the empirical CDF of each annotator's responses---with values closer to 0 implying lower confidence and those nearer 1 implying higher confidence. For questions that are dynamically revealed on the basis of the answer to the \textit{natural parts} question---i.e. \textit{part similarity}, \textit{average part duration}, \textit{dynamicity}, and \textit{situation duration}---we use the average of the ridit scored confidence for \textit{natural parts} and that question.

\begin{figure}[t]
    \centering
    \includegraphics[width=0.95\columnwidth]{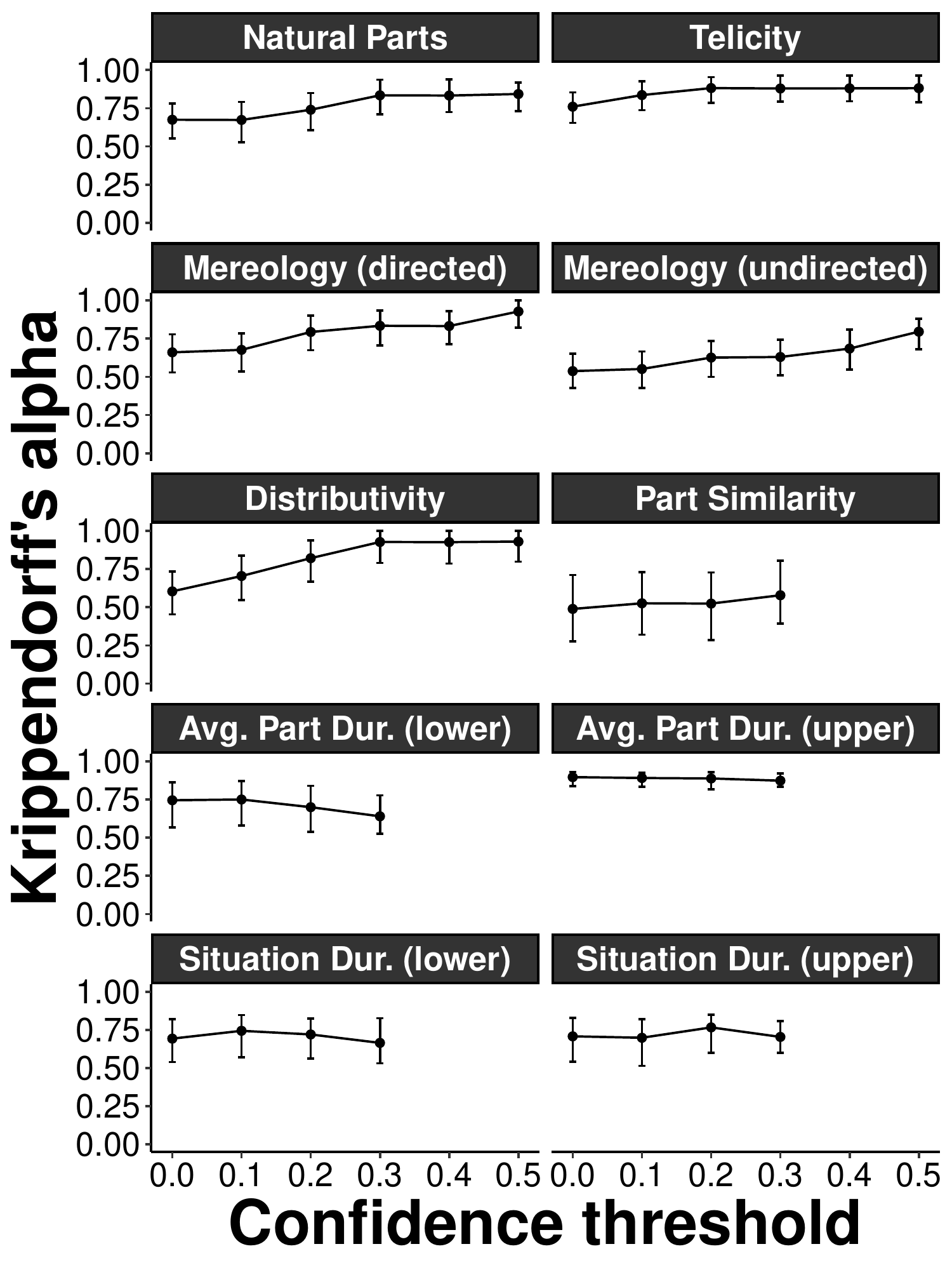}
    \vspace{-5mm}
    \caption{IAA among experts for each property, filtering annotations with ridit-scored confidence ratings below different thresholds. Confidence threshold 0.0 implies no filtering. Errors bars show 95\% confidence internals computed by a nonparametric bootstrap.}
    \label{fig:expert-agreement}
    \vspace{-5mm}
\end{figure}

\autoref{fig:expert-agreement} shows $\alpha$ when including only items that the expert annotators rated with a particular ridit scored confidence or higher. The agreement for the event-event protocol (mereology) is given in two forms: given that $e_1$ temporally contains $e_2$, (i) \textit{directed}: the agreement on whether $e_2$ is a subevent of $e_1$; and (ii) \textit{undirected}: the agreement on whether $e_2$ is a subevent of $e_1$ \textit{and} whether $e_1$ is a subevent of $e_2$. 

The error bars are computed by a nonparametric bootstrap over items. A threshold of 0.0 corresponds to computing $\alpha$ for all annotations, regardless of confidence; a threshold of $t >$ 0.0 corresponds to computing $\alpha$ only for annotations associated with a ridit scored confidence of greater than $t$. When this thresholding results in less than $\frac{1}{3}$ of items having an annotation for at least two annotators, $\alpha$ is not plotted. This situation occurs only for questions that are revealed based on the answer to a previous question. 

%
\begin{figure}[t]
    \centering
    \includegraphics[width=0.95\columnwidth]{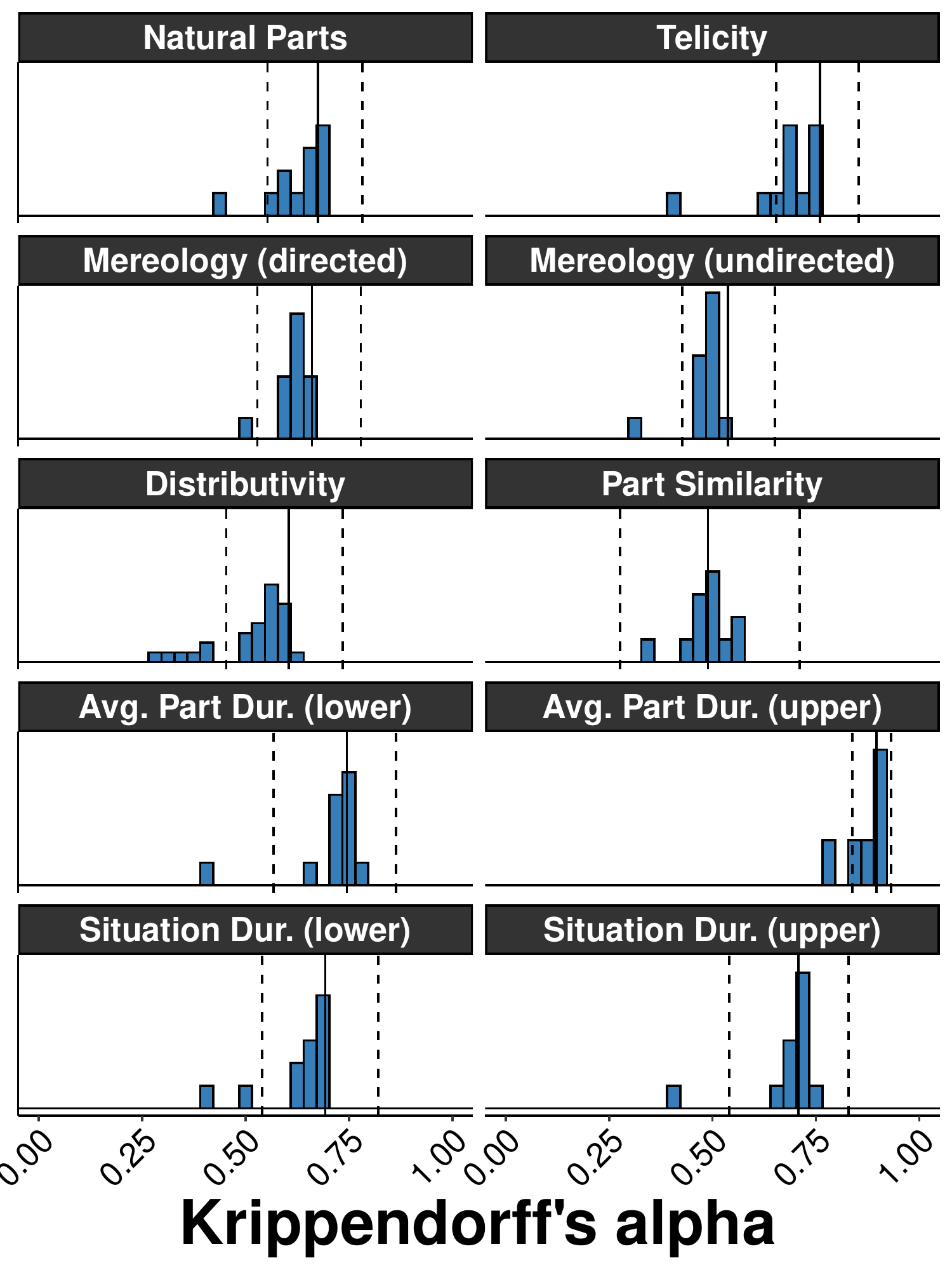}
    \vspace{-5mm}
    \caption{Per-property histograms of alphas for IAA between each crowd-sourced annotator and all experts. Black lines show the experts-only alpha, with dashed lines for the 95\% CI. (see Figure \ref{fig:expert-agreement}).}
    \label{fig:crowd-sourced-and-expert-agreement}
    \vspace{-5mm}
\end{figure}

For \textit{natural parts}, \textit{telicity}, \textit{mereology}, and \textit{distributivity}, agreement is high, even without filtering any responses on the basis of confidence, and that agreement improves with confidence. For \textit{part similarity}, \textit{average part duration}, and \textit{situation duration}, we see more middling, but still reasonable, agreement, though this agreement does not reliably increase with confidence. The fact that it does not increase may have to do with interactions between confidence on the \textit{natural parts} question and its dependent questions that we do not capture by taking the mean of these two confidences. 

\vspace{-2mm}

\paragraph{Crowd-Sourced Annotators}

We recruit crowd-sourced annotators in two stages. First, we select a small set of items from the 100 we annotate in the expert annotation that have high agreement among experts to create a qualification task. Second, based on performance in this qualification task, we construct a pool of trusted annotators who are allowed to participate in pilot annotations for each of the three subprotocols.\footnote{During all validation stages as well as bulk annotation (\S\ref{sec:corpus-annotation}), we targeted an average hourly pay equivalent to that for undergraduate research assistants performing corpus annotation at the first and final author's institution: \$12.50 per hour. A third-party estimate from TurkerView shows our actual average hourly compensation when data collection was completed to be \$14.71 per hour.}

\vspace{-2mm}

\subparagraph{Qualification}

For the qualification task, we selected eight of the sentences collected from MASC for the event-subevent subprotocol on which expert agreement was very high and which were diverse in the types of events described. We then obtained event-subevent annotations for these sentences from 400 workers on Amazon Mechanical Turk (AMT), and selected the top 200 among them on the basis of their agreement with expert responses on the same items. These workers were then permitted to participate in the pilot tasks. 


\vspace{-2mm}

\subparagraph{Pilot}

We conducted one pilot for each subprotocol, using the items described in \S\ref{ssec:item-selection}. Sentences were presented in lists of 10 per Human Intelligence Task (HIT) on AMT for the event-event and event-entity subprotocols and in lists of 5 per HIT for event-subevent. We collected annotations from 10 distinct workers for each sentence, and workers were permitted to annotate up to the full 100 sentences in each pilot. Thus, all pilots were guaranteed to include a minimum of 10 distinct workers (all workers do all HITs), up to a maximum of 100 for the subprotocols with 10 sentences per HIT or 200 for the subprotocol with 5 per HIT (each worker does one HIT). All top-200 workers from the qualification were allowed to participate. 

\autoref{fig:crowd-sourced-and-expert-agreement} shows IAA between all pilot annotators and experts for individual questions across the three pilots. More specifically, it shows the distribution of $\alpha$ scores by question for each annotator when IAA is computed among the three experts and that annotator only. Solid vertical lines show the expert-only IAA and dashed vertical lines show the 95\% confidence interval.



\subsection{Protocol Comparison}
\label{ssec:protocol-comparison}

To further validate the event-event and event-subevent subprotocols, we evaluate how well our pilot data predicts the corresponding \textsc{contains} v. \textsc{contains-subevent} annotations from RED in the former case, as well as the \textsc{event} v. \textsc{state} and \textsc{telic} v. \textsc{atelic} annotations from SitEnt in the latter. In both cases, we used the (ridit-scored) confidence-weighted average response across annotators for a particular item as features in a simple SVM classifier with linear kernel. In a leave-one-out cross-validation on the binary classification task for RED, we achieve a micro-averaged F1 score of 0.79---exceeding the reported human F1 agreement for both the \textsc{contains} (0.640) and \textsc{contains-subevent} (0.258) annotations reported by \citet{ogorman-etal-2016-richer}.

For SitEnt, we evaluate on a three-way classification task for \textsc{stative}, \textsc{eventive-telic}, and \textsc{eventive-atelic}, achieving a micro-averaged F1 of 0.68 using the same leave-one-out cross-validation. As \citet{friedrich_automatic_2014} do not report interannotator agreement for this class breakdown, we further compute Krippendorff's alpha from their raw annotations and again find that agreement between our predicted annotations and the gold ones (0.48) slightly exceeds the interannotator agreement among humans (0.47).

These results suggest that our subprotocols capture relevant event structural phenomena as well as linguistically trained annotators can and that they may serve as effective alternatives to existing protocols while not requiring any linguistic expertise.

\section{Corpus Annotation}
\label{sec:corpus-annotation}

We collect crowd-sourced annotations for the entirety of UD-EWT. Predicate and argument spans are obtained from the PredPatt predicate-argument graphs for UD-EWT available in UDS1.0. The total number of items annotated for each subprotocol is presented in \autoref{tab:corpus-examples}. 

\vspace{-2mm}

\paragraph{Event-subevent}
These annotations cover all predicates headed by verbs (as identified by UD POS tag), as well as copular constructions with nominal and adjectival complements. In the former case, only the verb token is highlighted in the task; in the latter, the highlighting  spans from the copula to the complement head.

\vspace{-2mm}

\paragraph{Event-event}
Pairs for the event-event subprotocol were drawn from the UDS-Time dataset, which features pairs of verbal predicates, either within the same sentence or in adjacent sentences, each annotated with its start- and endpoint relative to the other. We additionally included predicate-argument pairs in cases where the argument is annotated in UDS as having a WordNet supersense of \textsc{event}, \textsc{state}, or \textsc{process}. To our knowledge, this represents the largest publicly available (partial) event coreference dataset to date.

\vspace{-2mm}

\paragraph{Event-entity}
For the event-entity subprotocol, we identify predicate-argument pairs in which the argument is plural or conjoined. Plural arguments are identified by the UD \textsc{number} attribute, and conjoined ones by a \texttt{conj} dependency between an argument head and another noun. We consider only predicate-argument pairs with a UD dependency of \texttt{nsubj}, \texttt{nsubjpass}, \texttt{dobj}, or \texttt{iobj}.

\section{Event Structure Induction}
\label{sec:event-structure-induction}

Our goal in inducing event structural categories is to learn representations of those categories on the basis of annotated UDS graphs, augmented with the new UDS-E annotations. We aim to learn four sets of interdependent classifications grounded in UDS properties: event types, entity types, semantic role types, and event-event relation types. These classifications are interdependent in that we assume a generative model that incorporates both sentence- and document-level structure.\footnote{See \citealt{ferraro2016unified} for a related model that uses FrameNet's ontology, rather than inducing its own.}


\vspace{-2mm}
\paragraph{Document-level UDS}

Semantics edges in UDS1.0 represent only sentence-internal semantic relations. This constraint implies that annotations for cross-sentential semantic relations---a significant subset of our event-event annotations---cannot be represented in the graph structure. To remedy this, we extend UDS1.0 by adding \textit{document edges} that connect semantics nodes either within a sentence or in two distinct sentences, and we associate our event-event annotations with their corresponding document edge (see \autoref{fig:uds_semantics_graph}). Because UDS1.0 does not have a notion of document edge, it does not contain \citeauthor{vashishtha-etal-2019-fine}'s (\citeyear{vashishtha-etal-2019-fine}) fine-grained temporal relation annotations, which are highly relevant to event-event relations. We additionally add those attributes to their corresponding document edges.

\newcommand{\eventtypes}{\mathcal{T}_\mathrm{event}}
\newcommand{\entitytypes}{\mathcal{T}_\mathrm{ent}}
\newcommand{\relationtypes}{\mathcal{R}_\mathrm{rel}}
\newcommand{\roletypes}{\mathcal{R}_\mathrm{role}}
\newcommand{\documents}{\mathcal{D}}
\newcommand{\sentences}{\mathcal{S}}
\newcommand{\eventannotators}{\mathcal{A}^\mathrm{(event)}}
\newcommand{\entityannotators}{\mathcal{A}^\mathrm{(ent)}}
\newcommand{\roleannotators}{\mathcal{A}^\mathrm{(role)}}
\newcommand{\relationannotators}{\mathcal{A}^\mathrm{(rel)}}
\newcommand{\predargrelationannotators}{\mathcal{A}^\mathrm{(argrel)}}
\newcommand{\nodes}{\mathcal{V}}
\newcommand{\predicatenodes}{\mathcal{V}}
\newcommand{\argumentnodes}{\mathcal{A}}
\newcommand{\eventproperties}{\mathcal{P}_\mathrm{event}}
\newcommand{\entityproperties}{\mathcal{P}_\mathrm{ent}}
\newcommand{\roleproperties}{\mathcal{P}_\mathrm{role}}
\newcommand{\relationproperties}{\mathcal{P}_\mathrm{rel}}
\newcommand{\Bern}{\mathrm{Bern}}
\newcommand{\Cat}{\mathrm{Cat}}
\newcommand{\Normal}{\mathcal{N}}
\newcommand{\eonestart}{\overset{\leftarrow}{e_1}}
\newcommand{\etwostart}{\overset{\leftarrow}{e_2}}
\newcommand{\eoneend}{\overset{\rightarrow}{e_1}}
\newcommand{\etwoend}{\overset{\rightarrow}{e_2}}

\SetInd{0.1em}{0.5em}

\begin{algorithm}[t]
\scriptsize
Initialize queue $I$ \;
\For{sentence $s \in \sentences$}{
  Initialize queue $J$ \;
  Enqueue $J \rightarrow I$ \;
  \If{$\mathrm{length}(I) > W$}{
    Dequeue $I$
  }
  \For{predicate node $v \in \mathrm{predicates}(s)$}{
    Sample event type $t_{sv} \sim \Cat\left(\bm\theta^\mathrm{(event)}\right)$ \;
    \For{property $p \in \eventproperties$}{
      \For{annotator $i \in \eventannotators_{svp}$}{
        Sample $x^\mathrm{(event)}_{svpi} \sim f_p^i\left(\bm\mu^\mathrm{(event)}_{t_{sv}}\right)$
      }
    }
    Enqueue $\langle s, v \rangle \rightarrow J$\;
    \For{argument node $v' \in \mathrm{arguments}(s, v)$}{
      Sample ent. type $t_{sv'} \sim \Cat\left(\bm\theta^\mathrm{(entity)}\right)$ \;
      \For{property $p \in \entityproperties$}{
        \For{annotator $i \in \entityannotators_{sv'p}$}{
          Sample $x^\mathrm{(part)}_{sv'pi} \sim f_p^i\left(\bm\mu^\mathrm{(part)}_{t_{sv'}}\right)$ 
        }
      }
      \If{$v'$ is eventive}{
        Enqueue $\langle s, v' \rangle \rightarrow J$\;
      }
      Sample role type $r_{svv'} \sim \Cat\left(\bm\theta^\mathrm{(role)}_{t_{sv}t_{sv'}}\right)$ \;
      \For{property $p \in \roleproperties$}{
        \For{annotator $i \in \roleannotators_{svv'p}$}{
          Sample $x^\mathrm{(role)}_{svv'pi} \sim f_p^i\left(\bm\mu^\mathrm{(role)}_{r_{svv'}}\right)$
        }
      }
    }
    \For{index pair $\langle s', v' \rangle\in$ $\mathrm{flatten}(I)$}{
      Sample rel. type $q \sim \Cat\left(\bm\theta^\mathrm{(rel)}_{t_{sv}t_{s'v'}}\right)$ \;
      \For{property $p \in \relationproperties$}{
        \For{annotator $i \in \relationannotators_{svs'v'p}$}{
          Sample $x^\mathrm{(rel)}_{svs'v'pi} \sim f_p^i\left(\bm\mu^\mathrm{(rel)}_q\right)$
        }
      }
    }
  }
}
 \caption{Generative story of event structure induction model for a single document with sentence window $W$\vspace{-6mm}}
 \label{alg:generative-story}
\end{algorithm}

\vspace{-2mm}
\paragraph{Generative Model}

\begin{figure*}
    \centering
    \includegraphics[width=1.7\columnwidth]{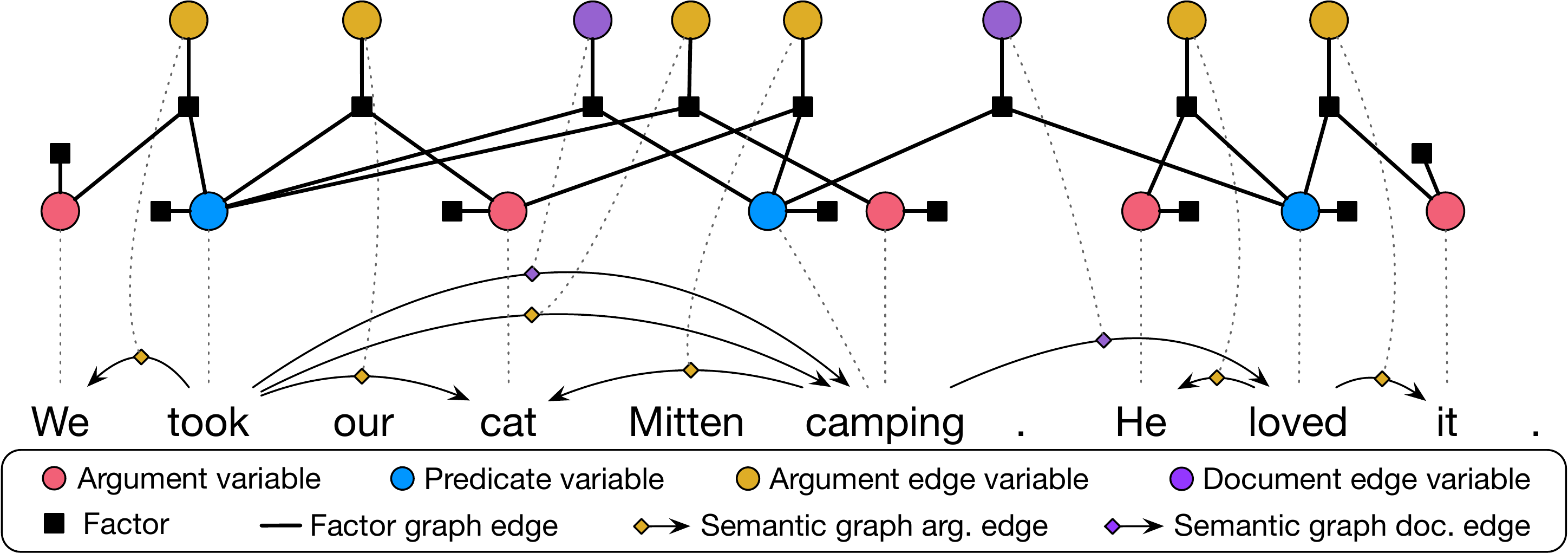}
    \vspace{-3mm}
    \caption{The factor graph for the pair of sentences shown in \autoref{fig:uds_semantics_graph} based on the generative story given in \autoref{alg:generative-story}. Each node or edge annotated in the semantics graphs becomes a variable node in the factor graph, as indicated by the dotted lines. Only factors for the prior distributions over types are shown; the annotation likelihood factors associated with each variable node are omitted for space.}
    \vspace{-5mm}
    \label{fig:factor_graph}
\end{figure*}

Algorithm \ref{alg:generative-story} gives the generative story for our event structure induction model. We assume some number of types of events $\eventtypes$, roles $\roletypes$, entities $\entitytypes$, and relations $\relationtypes$. \autoref{fig:factor_graph} shows the resulting factor graph for the semantic graphs shown in \autoref{fig:uds_semantics_graph}.

\vspace{-1mm}

\subparagraph{Annotation Likelihoods}
The distribution $f_p^a$ on the annotations themselves is implemented as a mixed model \citep{gelman_data_2014} dependent on property $p$ being annotated with annotator random intercepts $\bm{R}$, where the random intercepts for annotator $a$ are $\bm\rho_{a} \sim \Normal(\bm{0}, \bm\Sigma_\mathrm{ann})$ with unknown $\bm\Sigma_\mathrm{ann}$. When $p$ receives binary annotations, a simple logistic mixed model is assumed, where $f_p^a = \Bern(\mathrm{logit}^{-1}(\mu_{i_p} + \rho_{ai_p}))$ and $i_p$ is the index corresponding to property $p$ in the expected annotation $\bm\mu$. When $p$ receives nominal annotations, $f_p^a = \Cat(\mathrm{softmax}(\bm \mu_{i_p} + \bm\rho_{ai_p}))$ and $i_p$ is a set with cardinality of the number of nominal categories. And when $p$ receives ordinal annotations, we follow \citet{white-etal-2020-universal} in using an ordinal (linked logit) mixed effects model where $\bm\rho_{a}$ defines the cutpoints between response values in the cumulative density function for annotator $a$:

\vspace{-7mm}

\begin{align*}
\mathbb{P}(x_{ai_p} \leq j) &= \mathrm{logit}^{-1}(\mu_{i_p} - \rho_{ai_p})\\
f_p^a(x_{ai_p} = j) &= \mathbb{P}(x_{ai_p} \leq j) - \mathbb{P}(x_{ai_p} \leq j-1)
\end{align*}

\vspace{-3mm}

\subparagraph{Conditional Properties}
For both our dataset and UDS-Protoroles, certain annotations are conditioned on others, owing to the fact that whether some questions are asked at all depends upon annotator responses to previous ones. Following \citet{white-etal-2017-semantic}, we model the likelihoods for these properties using hurdle models \citep{agresti_categorical_2014}: for a given property, a Bernoulli distribution determines whether the property applies; if it does, the property value is determined using a second distribution of the appropriate type.

\vspace{-1mm}

\subparagraph{Temporal Relations}
Temporal relations annotations from UDS-Time consist of 4-tuples $(\eonestart, \etwostart, \eoneend, \etwoend)$ of real values on the unit interval, representing start- and endpoints of two event-referring predicates or arguments, $e_1$ and $e_2$. Each tuple is normalized such that the earlier of $(\eonestart, \etwostart)$ is always locked to the left end of the scale (0) and the later of $(\eoneend, \etwoend)$ to the right end (1). The likelihood for these annotations must consider the different possible orderings of the two events. To do so, we first determine whether $\eonestart$ is locked, $\etwostart$ is, or both are, according to $\Cat\left(\mathrm{softmax}(\bm\mu_{\mathrm{lock^{\leftarrow}}} + \bm\rho_{ai_{\mathrm{lock^{\leftarrow}}}})\right)$. We do likewise for $\eoneend$ and $\etwoend$, using a separate distribution $\Cat\left(\mathrm{softmax}(\bm \mu_{\mathrm{lock^{\rightarrow}}} + \bm\rho_{ai_{\mathrm{lock^{\rightarrow}}}})\right)$. Finally, if the start point from one event and the endpoint from the other are free (i.e. not locked), we determine their relative ordering using a third distribution $\Cat\left(\mathrm{softmax}(\bm \mu_{\mathrm{lock^{\leftrightarrow}}} + \bm\rho_{ai_{\mathrm{lock^{\leftrightarrow}}}})\right)$.

\vspace{-2mm}
\paragraph{Implementation}

We fit our model to the training data using expectation-maximization. We use loopy belief propagation to obtain the posteriors over event, entity, role, and relation types in the expectation step and the Adam optimizer to estimate the parameters of the distributions associated with each type in the maximization step.\footnote{The variable and factor nodes for the relation types can introduce cycles into the factor graph for a document, which is what necessitates the loopy variant of belief propagation.} As a stopping criterion, we compute the evidence that the model assigns to the development data, stopping when this quantity begins to decrease.

To make use of the (ridit-scored) confidence response $c_{ai_p} \in (0, 1)$ associated with each annotation $x_{ai_p}$, we weight the log-likelihood of $x_{ai_p}$ by $c_{ai_p}$ when computing the evidence of the annotations. This weighting encourages the model to explain annotations that an annotator was highly confident in, penalizing the model less if it assigns low likelihood to a low confidence annotation.

To select $|\eventtypes|$, $|\entitytypes|$, $|\roletypes|$, and $|\relationtypes|$ for Algorithm \ref{alg:generative-story}, we fit separate mixture models for each classification---i.e. removing all factor nodes---using the same likelihood functions $f^i_p$ as in Algorithm \ref{alg:generative-story}. We then compute the evidence that the simplified model assigns to the development data given some number of types, choosing the smallest number such that there is no reliable increase in the evidence for any larger number. To determine reliability, we compute 95\% confidence intervals using nonparametric bootstraps. Importantly, this simplified model is only used to select $|\eventtypes|$, $|\entitytypes|$, $|\roletypes|$, and $|\relationtypes|$: all analyses below are conducted on full model fits.


\vspace{-2mm}
\paragraph{Types}

The selection procedure described above yields $|\eventtypes| = 4$, $|\entitytypes| = 8$, $|\roletypes| = 2$, and $|\relationtypes| = 5$. To interpret these classes, we inspect the property means $\bm\mu_t$ associated with each type $t$ and give examples from UD-EWT for which the posterior probability of that type is high.

\begin{figure}[t]
  \centering
     \includegraphics[width=0.85\columnwidth]{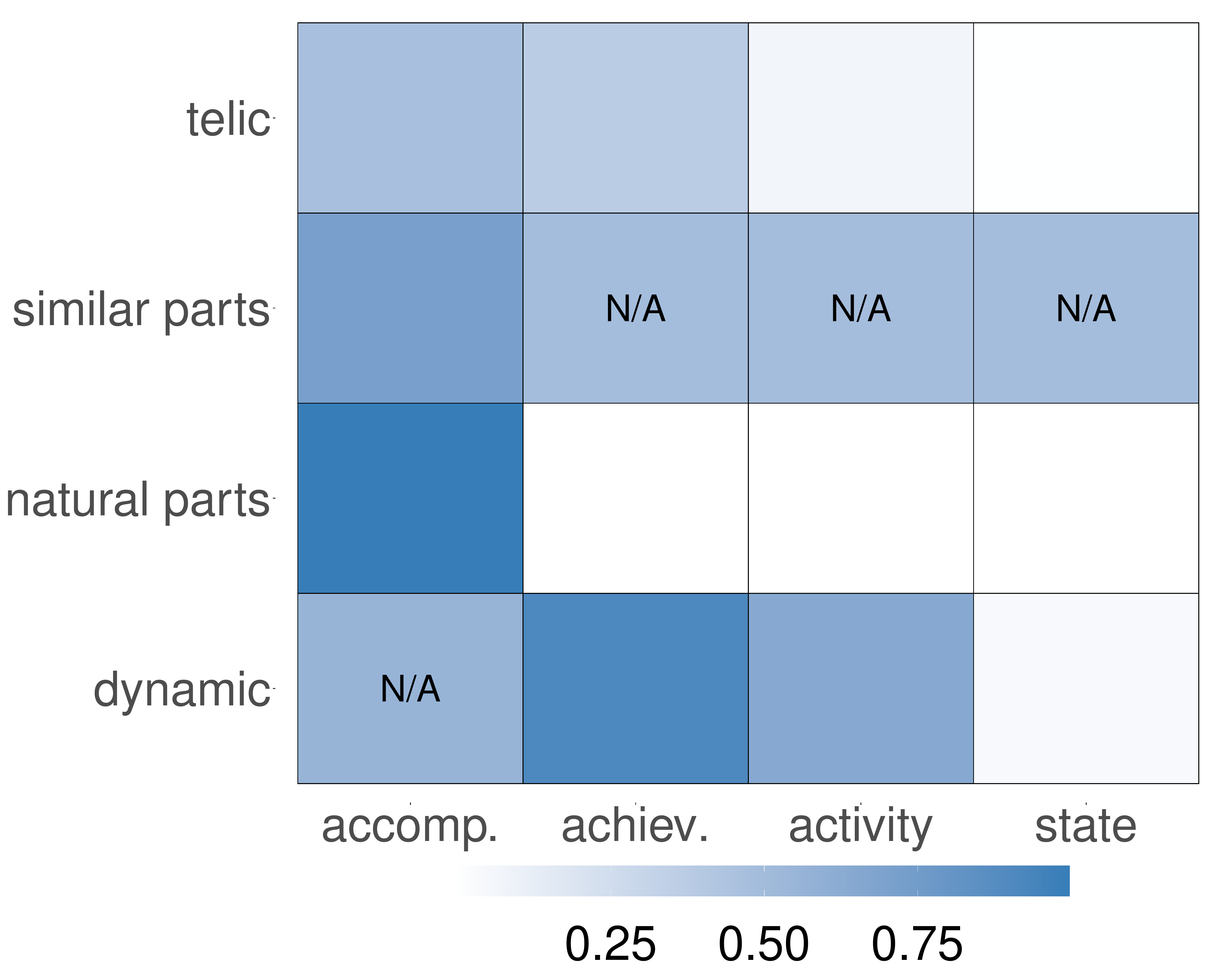}
    \vspace{-5mm}
     \caption{Probability of binary properties from the event-subevent protocol by event type. Cells marked with ``N/A'' indicate that the property generally does not apply for the corresponding type because of the conditional dependence on natural parts.}
     \label{fig:event_types}
     \vspace{-5mm}
 \end{figure}

\vspace{-3mm}

\subparagraph{Event Types} While our goal was not necessarily to reconstruct any particular classification from the theoretical literature, the four event types align fairly well with those proposed by \citet{vendler_verbs_1957}: statives \ref{ex:state-corpus}, activities \ref{ex:activity-corpus}, achievements \ref{ex:achievement-corpus}, and accomplishments  \ref{ex:accomplishment-corpus}. We label our clusters based on these interpretations (\autoref{fig:event_types}).

\ex. I have finally found a mechanic I \textbf{trust}!! \label{ex:state-corpus}

\ex. his agency is still \textbf{reviewing} the decision. \label{ex:activity-corpus}

\ex. A suit against [\ldots] Kristof was \textbf{dismissed}.  \label{ex:achievement-corpus}

\ex. a consortium [\ldots] \textbf{established} in 1997  \label{ex:accomplishment-corpus}

One difference between \citeauthor{vendler_verbs_1957}'s classes and our own is that our ``activities'' correspond primarily to those without dynamic subevents, while our ``accomplishments'' encompass both his accomplishments and activities with dynamic subevents (see discussion of \citealt{taylor_tense_1977} in \S\ref{sec:background}).

\begin{figure}
    \centering
    \includegraphics[width=.85\columnwidth]{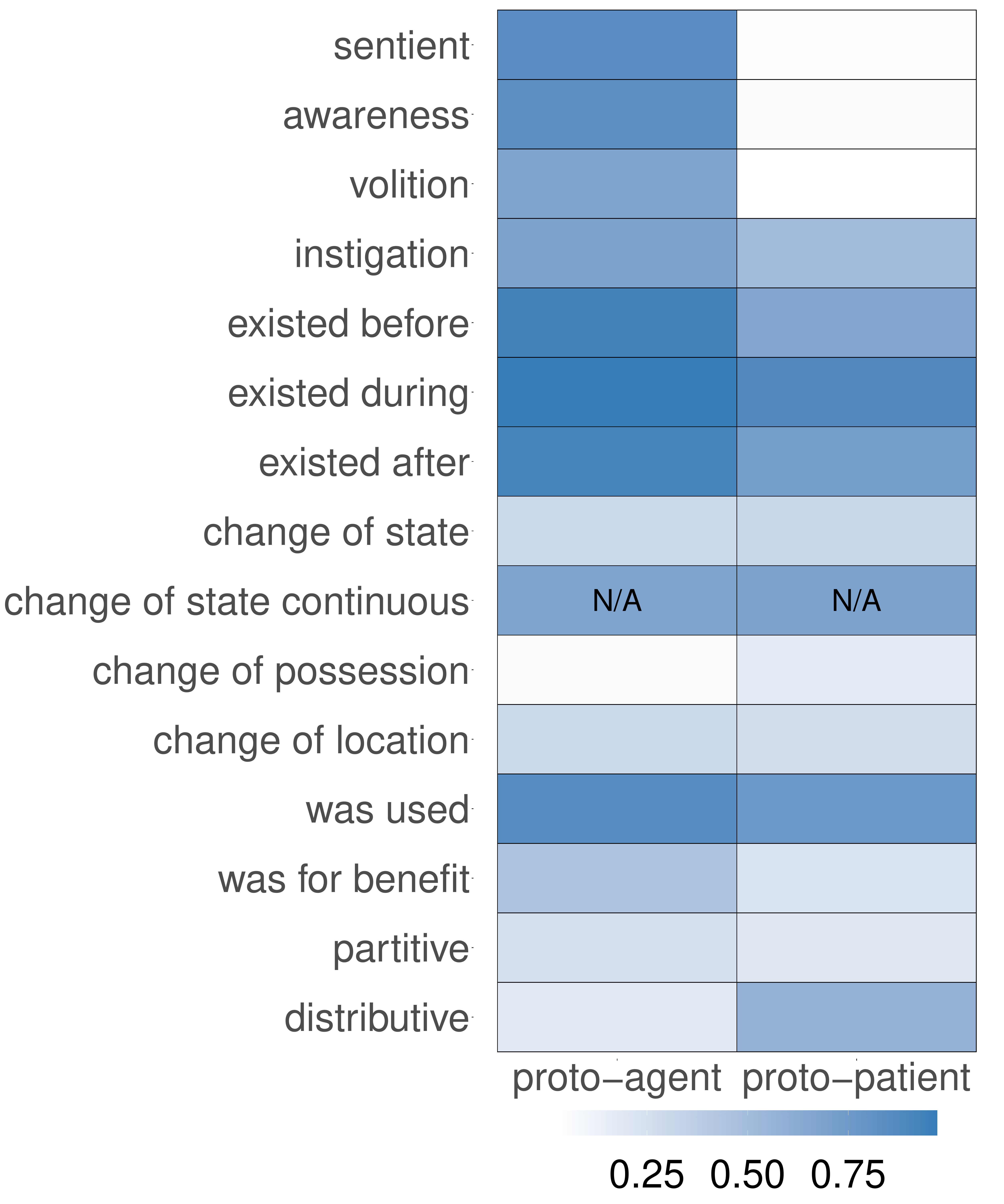}
    \vspace{-5mm}
    \caption{Probability of role type properties. These include existing UDS protoroles properties, along with the \textit{distributive} property from the event-participant subprotocol. We have labeled the role types with our proto-agent/proto-patient interpretation given below. \vspace{-5mm}}
    \label{fig:role_types}
\end{figure}

Even if approximate, this alignment is surprising given that \citeauthor{vendler_verbs_1957}'s classification was not developed with actual language use in mind and thus abstracts away from complexities that arise when dealing with, e.g., non-factual or generic events. Nonetheless, there do arise cases where a particular predicate has a wider distribution across types than we might expect based on prior work. For instance, \textit{know} is prototypically stative; and while it does get classed that way by our model, it also gets classed as an accomplishment or achievement (though rarely an activity)---e.g. when it is used to talk about coming to know something, as in \ref{ex:come-to-know}. 

\ex. Please let me \textbf{know} how[\ldots]to proceed. \label{ex:come-to-know}

\vspace{-3mm}
\subparagraph{Entity Types} Our entity types are: person/group \ref{ex:entity-people}, concrete artifact \ref{ex:entity-concrete-artifacts}, contentful artifact \ref{ex:entity-particular-contentful-artifacts}, particular state/event \ref{ex:entity-particular-cognitive-state}, generic state/event \ref{ex:entity-generic-cognitive-state}, time \ref{ex:entity-times}, kind of concrete objects \ref{ex:entity-other-kind}, and particular concrete objects \ref{ex:entity-other-concrete}.

\ex. Have a real \textbf{mechanic} check[...]\label{ex:entity-people}

\ex. I have a [\ldots] cockatiel, and there are 2 \textbf{eggs} in the bottom of the cage[\ldots] \label{ex:entity-concrete-artifacts}

\ex. Please find attached a credit \textbf{worksheet}[\ldots]\label{ex:entity-particular-contentful-artifacts}

\ex. He didn't take a \textbf{dislike} to the kids[...]\label{ex:entity-particular-cognitive-state}

\ex. They require \textbf{a lot of attention} [\ldots]\label{ex:entity-generic-cognitive-state}

\ex. Every move Google makes brings this particular \textbf{future} closer. \label{ex:entity-times}

\ex. And what is their big / main \textbf{meal} of the day.\label{ex:entity-other-kind}

\ex. Find him before he finds the dog \textbf{food}.\label{ex:entity-other-concrete}

\vspace{-3mm}
\subparagraph{Role Types} The optimality of two role types is consistent with  \citeauthor{dowty_thematic_1991}'s (\citeyear{dowty_thematic_1991}) proposal that there are only two abstract role prototypes---\textit{proto-agent} and \textit{proto-patient}---into which individual thematic roles---i.e. those specific to particular predicates---cluster. Further, the means for the two role types we find very closely track those predicted by \citeauthor{dowty_thematic_1991}, with clear proto-agents \ref{ex:proto-agent} and proto-patients \ref{ex:proto-patient} (see also \citealt{white-etal-2017-semantic}).

\ex.  \textbf{\textit{they}} don't \textbf{press} their sandwiches. \label{ex:proto-agent}

\ex. you don't ever feel like you \textbf{ate} too \textbf{\textit{much}}. \label{ex:proto-patient}

\vspace{-2mm}
\subparagraph{Relation Types} The relation types we obtain track closely with approaches that use sets of underspecified temporal relations \citep{cassidy-etal-2014-annotation,ogorman-etal-2016-richer,zhou-etal-2019-going,zhou-etal-2020-temporal,wang-etal-2020-joint}: $e_1$ starts before $e_2$ \ref{ex:e1-starts-first}, $e_2$ starts before $e_1$ \ref{ex:e2-starts-first}, $e_2$ ends after $e_1$ \ref{ex:e2-ends-last}, $e_1$ contains $e_2$ \ref{ex:containment}, and $e_1 = e_2$ \ref{ex:identity}.

\ex. [\ldots]the Spanish, Thai and other contingents are already \textbf{committed} to \textbf{leaving} [\ldots] \label{ex:e1-starts-first}

\ex. And I have to \textbf{wonder}: Did he \textbf{forget} that he already has a memoir[...]\label{ex:e2-starts-first} 

\ex. no, i am not \textbf{kidding} and no i don't want it b/c of the taco bell dog. i want it b/c it is really \textbf{small} and cute.\label{ex:e2-ends-last}

\ex.  they \textbf{offer} cheap air tickets to their country  [...] you may get excellent discount airfare, which may even \textbf{surprise} you. \label{ex:containment}

\ex. the food is good, however the tables are so \textbf{close together} that it feels very \textbf{cramped}. \label{ex:identity}

\vspace{-2mm}

\paragraph{Type consistency}

\begin{figure}
    \centering
    \includegraphics[width=\columnwidth]{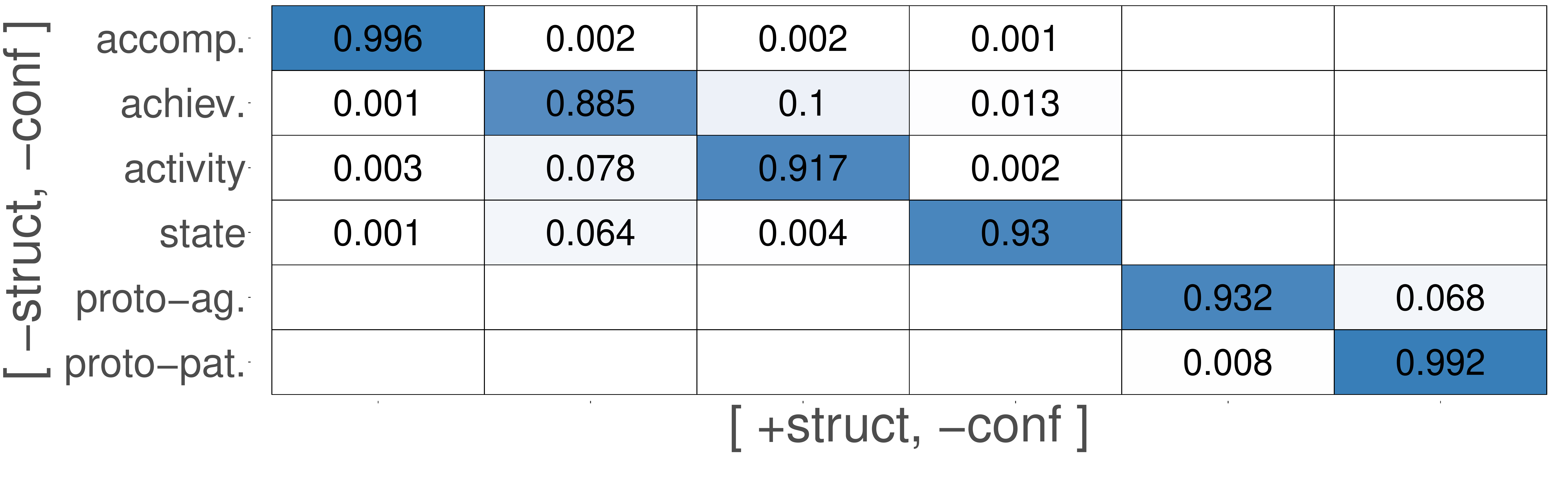}
    \includegraphics[width=\columnwidth]{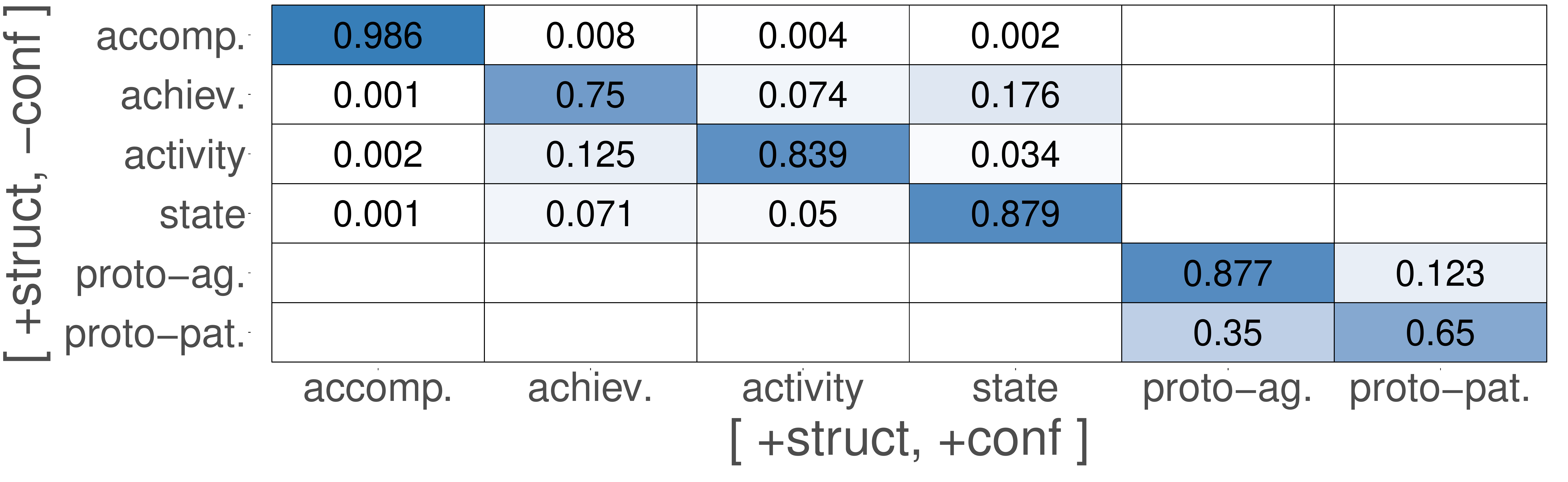}
    \vspace{-11mm}
    \caption{Confusion matrices for event and role types.}
    \label{fig:confusion_matrices}
    \vspace{-5mm}
\end{figure}

To assess the impacts of sentence/document-level structure (\autoref{alg:generative-story}) and confidence weighting on the types we induce, we investigate how the distributions over role and event types shift when comparing models fit with and without structure and confidence weighting. \autoref{fig:confusion_matrices} visualizes these shifts as row-normalized confusion matrices computed from the posteriors across items derived from each model. It compares (top) the simplified model used for model selection (rows) against the full model without confidence weighting (columns), and (bottom) the full model without confidence weighting (rows) against the one with (columns).\footnote{The distributional shifts for entity and relation types were extremely small, and so we do not discuss them here.} 

First, we find that the interaction between types afforded by incorporating the full graph structure (top plot) produces small shifts in both the event and role type distributions, suggesting that the added structure may help chiefly in resolving boundary cases, which is exactly what we might hope additional model structure would do. Second, weighting likelihoods by annotator confidence (bottom) yields somewhat larger shifts as well as more entropic posteriors (0.22 average normalized entropy for events; 0.30 for roles) than without weighting (0.02 for events; 0.22 for roles). 

Higher entropy is expected (and to some extent, desirable) here: introducing a notion of confidence should make the model less confident about items that annotators were less confident about. Further, among event types, the distribution of posterior entropy across items is driven by a minority of high uncertainty items, as evidenced by a very low median normalized entropy for event types (0.02). The opposite appears to be be true among the role types, for which the median is high (0.60). This latter pattern is perhaps not surprising in light of theoretical accounts of semantic roles, such as Dowty's: the entire point of such accounts is that it is very difficult to determine sharp role categories, suggesting the need for a more continuous notion. 


\section{Comparison to Existing Ontologies}
\label{sec:comparison-to-existing-ontologies}

To explore the relationship between our induced classification and existing event and role ontologies, we ask how well our event, role, and entity types map onto those found in PropBank and VerbNet. Importantly, the goal here is not perfect alignment between our types and PropBank and VerbNet types, but rather to compare other classifications that reflect top-down assumptions to the one we derive bottom-up.

\vspace{-2mm}

\paragraph{Implementation}

To carry out these comparisons, we use the parameters of the posterior distributions over event types $\bm\theta^\text{(ev)}_p$ for each predicate $p$, over role types $\bm\theta^\text{(role)}_{pa}$ for each argument $a$ of each predicate $p$, and over entity types $\bm\theta^\text{(ent)}_{pa}$ for each argument $a$ of each predicate $p$ as features in an SVM with RBF kernel predicting the event and role types found in PropBank and VerbNet. We take this route, over direct comparison of types, to account for the possibility that information encoded in role or event types within VerbNet or PropBank is distributed differently in our more abstract classification. We tune L2 regularization ($\lambda \in$ \{1, 0.5, 0.2, 0.1, 0.01, 0.001\}) and bandwidth ($\gamma \in$ \{1e-2, 1e-3, 1e-4, 1e-5\}) using grid search, selecting the best model based on performance on the standard UD-EWT development set. All metrics reflect UD-EWT test set performance.

\vspace{-2mm}

\paragraph{Role Type Comparison}

\begin{table}
\footnotesize
\centering
\begin{tabular}{cccccc}
\toprule
          & \textbf{Role} &  \textbf{P} &  \textbf{R} &    \textbf{F} & \textbf{Micro F} \\
\midrule
   \multirow{2}{*}{argnum} &        A0 &       0.58 &    0.63 &  0.60 & \multirow{2}{*}{0.67} \\
    &        A1 &       0.72 &    0.78 &  0.75 & \\
    \midrule
  \multirow{2}{*}{functag} &      pag &       0.57 &    0.59 &  0.58 & \multirow{2}{*}{0.62} \\
   &      ppt &       0.65 &    0.77 &  0.71 & \\
   \midrule
  \multirow{3}{*}{verbnet}  &    agent &       0.64 &    0.54 &  0.59 & \multirow{3}{*}{NA}  \\
   &  patient &       0.20 &    0.14 &  0.16 & \\
   &    theme &       0.55 &    0.58 &  0.57 & \\
\bottomrule
\end{tabular}
\vspace{-2mm}
\caption{Test set results for all role types that are labeled on at least 5\% of the development data.}
\label{tab:role-comparison}
\vspace{-5mm}
\end{table}

We first obtain a mapping from UDS predicates and arguments to the PropBank predicates and arguments annotated in EWT. Each such argument in PropBank is annotated with an argument number (\textsc{A0}-\textsc{A4}) as well as a function tag (\textsc{pag} = \textit{agent}, \textsc{ppt} = \textit{patient}, etc.). We then compose this mapping with the mapping given in the PropBank frame files from PropBank rolesets to sets of VerbNet classes and from PropBank roles to sets of VerbNet roles (\textsc{agent}, \textsc{patient}, \textsc{theme}, etc.) to obtain a mapping from UDS arguments to sets of VerbNet roles. Because a particular argument maps to a set of VerbNet roles, we treat predicting VerbNet roles as a multi-label problem, fitting one SVM per role. For each argument $a$ of predicate $p$, we use as predictors $[\bm\theta^\text{(ev)}_p; \bm\theta^\text{(role)}_{pa}; \bm\theta^\text{(ent)}_{pa}; \bm\theta^\text{(role)}_{p\lnot a}; \bm\theta^\text{(ent)}_{p\lnot a}]$, with $\theta^\text{(role/ent)}_{p\lnot aj} = [\max_{a'\neq a} \theta^\text{(role/ent)}_{pa'j}, \mathrm{mean}_{a'\neq a} \theta^\text{(role/ent)}_{pa'j}]$.


\autoref{tab:role-comparison} gives the test set results for all role types labeled on at least 5\% of the development data. For comparison, a majority guessing baseline obtains micro F1s of 0.58 (argnum) and 0.53 (functag).\footnote{A majority baseline for VerbNet roles always yields an F1 of 0 in our multi-label setup, since no role is assigned to more than half of arguments.} Our roles tend to align well with agentive roles---\textsc{pag}, \textsc{agent}, and A0---and some non-agentive roles---\textsc{ppt}, \textsc{theme}, and A1---but they align less well with other non-agentive roles---\textsc{patient}. This result suggests that our two-role classification aligns fairly closely with the agentivity distinctions in PropBank and VerbNet, as we would expect if our roles in fact captured something like \citeauthor{dowty_thematic_1991}'s coarse distinction among prototypical agents and patients.

\vspace{-2mm}
\paragraph{Event Type Comparison}


The PropBank roleset and VerbNet class ontologies are extremely fine-grained, with PropBank capturing specific predicate senses and VerbNet capturing very fine-grained syntactic behavior of a generally small set of predicates. Since our event types are intended to be more general than either, we do not compare it directly to PropBank rolesets or VerbNet classes.

Instead, we compare to the generative lexicon-inspired variant of VerbNet's semantics layer \citep{brown-etal-2018-integrating}.  An example of this layer for the predicate  \texttt{give-13.1} is \textsf{\small has\_possession(e1, Ag, Th) \& transfer(e2, Ag, Th, Rec) \& cause(e2, e3) \& has\_possession(e3, Rec, Th)}. We predict only the abstract predicates in this decomposition---e.g. \textsf{transfer} or \textsf{cause}---treating the problem as multi-label and fitting one SVM per predicate. For each predicate $p$, we use as predictors $[\bm\theta^\text{(ev)}_p; \bm\theta^\text{(role)}_{p\cdot}; \bm\theta^\text{(ent)}_{p\cdot}]$, with $\theta^\text{(role/ent)}_{p \cdot j} = [\max_a \theta^\text{(role/ent)}_{paj}, \mathrm{mean}_a \theta^\text{(role/ent)}_{paj}]$.

\autoref{tab:event-comparison} gives the test set results for the five most frequent predicates in the corpus. For comparison, a majority guessing baseline would yield the same F (0.66) as our model for \textsc{cause}, but since none of the other classes are assigned to more than half of events, majority guessing for those would yield an F of 0. This result suggests that, while there may be some agreement between our classification and VerbNet's semantics layer, the two representations are relatively distinct.

\begin{table}
\small
\centering
\begin{tabular}{ccccc}
\toprule
         \textbf{Predicate} &  \textbf{P} &  \textbf{R} &    \textbf{F} \\
\midrule
          cause  &       0.51 &    0.95 &  0.66 \\
             do  &       0.30 &    0.25 &  0.27 \\
 has\_possession &       0.23 &    0.18 &  0.20 \\
   has\_location &       0.11 &    0.14 &  0.12 \\
         motion  &       0.09 &    0.10 &  0.09 \\
\bottomrule
\end{tabular}
\vspace{-2mm}
\caption{Test set results for all VerbNet predicates that are labeled on five most frequent predicates.}
\label{tab:event-comparison}
\vspace{-5mm}
\end{table}

\section{Conclusion}
\label{sec:conclusion}

We have presented an event structure classification derived from inferential properties annotated on sentence- and document-level semantic graphs. We induced this classification jointly with semantic role, entity, and event-event relation types using a document-level generative model. Our model identifies types that approximate theoretical predictions---notably, four event types like Vendler's, as well as proto-agent and proto-patient role types like Dowty's. We hope this work encourages greater interest in computational approaches to event structural understanding while also supporting work on adjacent problems in NLU, such as temporal information extraction and (partial) event coreference, for which we provide the largest publicly available dataset to date.


\section*{Acknowledgments}
\label{sec:acknowledgments}

We would like to thank Emily Bender, Dan Gildea, and three anonymous reviewers for detailed comments on this paper. We would also like to thank members of the Formal and Computational Semantics lab at the University of Rochester for feedback on the annotation protocols. This work was supported in part by the National Science Foundation (BCS-2040820/2040831, \textit{Collaborative Research: Computational Modeling of the Internal Structure of Events}) as well as by DARPA AIDA and DARPA KAIROS. The views and conclusions contained in this work are those of the authors and should not be interpreted as necessarily representing the official policies, either expressed or implied, or endorsements of DARPA or the U.S. Government. The U.S. Government is authorized to reproduce and distribute reprints for governmental purposes notwithstanding any copyright annotation therein.

\bibliography{anthology,references,extra}
\bibliographystyle{acl_natbib}

\end{document}